\documentclass[conference]{IEEEtran}
\IEEEoverridecommandlockouts
\usepackage{amsmath,amssymb,amsfonts}
\usepackage{algorithmic}
\usepackage{graphicx}
\usepackage{textcomp}
\usepackage{xcolor}

\usepackage{caption}
\usepackage{url}
\usepackage{algorithm}
\usepackage{algorithmic}
\usepackage{graphicx}
\usepackage{amsmath}
\usepackage{amsfonts}
\usepackage{booktabs}
\usepackage{xcolor}
\usepackage{mathrsfs}
\usepackage{arydshln}
\usepackage{multicol}
\usepackage{times}
\usepackage{multirow}
\usepackage{latexsym}
\usepackage{microtype}
\usepackage[square,sort,comma,numbers]{natbib}
\usepackage{times}
\usepackage{latexsym}
\usepackage{microtype}

\def\BibTeX{{\rm B\kern-.05em{\sc i\kern-.025em b}\kern-.08em
    T\kern-.1667em\lower.7ex\hbox{E}\kern-.125emX}}

\begin{document}

\title{Diversity-Driven Combination \\ for Grammatical Error Correction
}

\author{\IEEEauthorblockN{Wenjuan Han}
\IEEEauthorblockA{\textit{Department of Computer Science} \\
\textit{National University of Singapore}\\
hanwj0309@gmail.com
}
\and
\IEEEauthorblockN{Hwee Tou Ng}
\IEEEauthorblockA{\textit{Department of Computer Science} \\
\textit{National University of Singapore}\\
nght@comp.nus.edu.sg}
}

\maketitle

\begin{abstract}

Grammatical error correction (GEC) is the task of detecting and correcting errors in a written text. The idea of combining multiple system outputs has been successfully used in GEC. To achieve successful system combination, multiple component systems need to produce corrected sentences that are both diverse and of comparable quality. However, most existing state-of-the-art GEC approaches are based on similar sequence-to-sequence neural networks, so the gains are limited from combining the outputs of component systems similar to one another. In this paper, we present Diversity-Driven Combination (DDC) for GEC, a system combination strategy that encourages diversity among component systems. We evaluate our system combination strategy on the CoNLL-2014 shared task and the BEA-2019 shared task. On both benchmarks, DDC achieves significant performance gain with a small number of training examples and outperforms the component systems by a large margin. Our source code is available at \url{https://github.com/nusnlp/gec-ddc}.

\end{abstract}

\maketitle

\section{Introduction}

Grammatical error correction (GEC) is the task of detecting and correcting errors in a written text. A GEC system benefits a second language learner as a learning aid by taking a potentially erroneous sentence as input and transforming it to its corrected version. 

GEC increasingly gains significant attention since the HOO \cite{dale2010helping} and CoNLL shared tasks \cite{ng2013conll,ng2014conll}. Different approaches to high-quality grammatical error correction have been proposed. The idea of system combination -- combining corrected sentences predicted by several existing component systems -- has been successfully used in GEC \cite{susanto2014system,kantor2019learning,lin2021sc}. To achieve successful system combination, multiple component systems need to produce corrected sentences that are both diverse and of comparable quality \cite{macherey2007empirical}.
However, most existing state-of-the-art GEC approaches are based on similar sequence-to-sequence neural networks, especially transformer-based architectures \cite{kiyono2019empirical,kantor2019learning,omelianchuk2020gector}. This leads to a challenge for system combination since combining multiple system outputs benefits from differences between corrected sentences. The gains are limited when combining outputs from component systems similar to each other. In the extreme case, if the component systems are identical, the combination of these components will not bring any benefits.
In this work, we present Diversity-Driven Combination (DDC) for GEC, a system combination strategy for text generation tasks to take advantage of component systems. DDC consists of two parts: the component systems with a trainable component as the backbone, and the combination system. Before combining, the backbone component is trained by reinforcement learning based on a designed reward function that rewards the backbone component for producing corrected sentences which are different from an ensemble of outputs predicted by other component systems.
During combination, we adopt the system combination technique of Heafifield and Lavie \cite{heafield2010combining} to combine the outputs from the component systems to produce better results. The contributions of our paper are as follows.

\begin{itemize}

   \item We propose a novel system combination strategy for text generation tasks to increase cross-system diversity while maintaining the performance of each component system. By combining the outputs of diverse component systems,  the strengths of some will offset the weaknesses of the others and the final output can take advantage of all component systems.

    \item We propose a way to combine component systems as a black box, except for the backbone component. It relies solely on the outputs for non-backbone components and does not need to know the internal details of the non-backbone components (such as model structure, optimization strategy, hyperparameters, training dataset, and gradients over each layer).
        
     \item The lack of in-domain labeled data for training is a bottleneck for GEC. For our proposed method, less than five thousand training samples are used to train the combination system and the backbone component towards diversity. Compared to training single models that is computationally expensive and memory intensive (e.g., transformer-based seq2seq model \cite{choe2019neural}), our strategy is much more efficient.    % 4384
     
     \item Our proposed strategy can be easily applied to other scenarios and tasks. We only require one of the component systems to be trainable.
     
\end{itemize}

We evaluate our system combination strategy on the BEA-2019 shared task \cite{bryant2019bea} and the CoNLL-2014 shared task \cite{ng2014conll}. On both benchmarks, DDC achieves significant performance gain with a small number of training samples.

\section{Diversity-Driven Combination}
      \begin{figure*}[]
      \begin{center}
      \includegraphics[width=1.55\columnwidth,trim=0 0 0 2,clip]{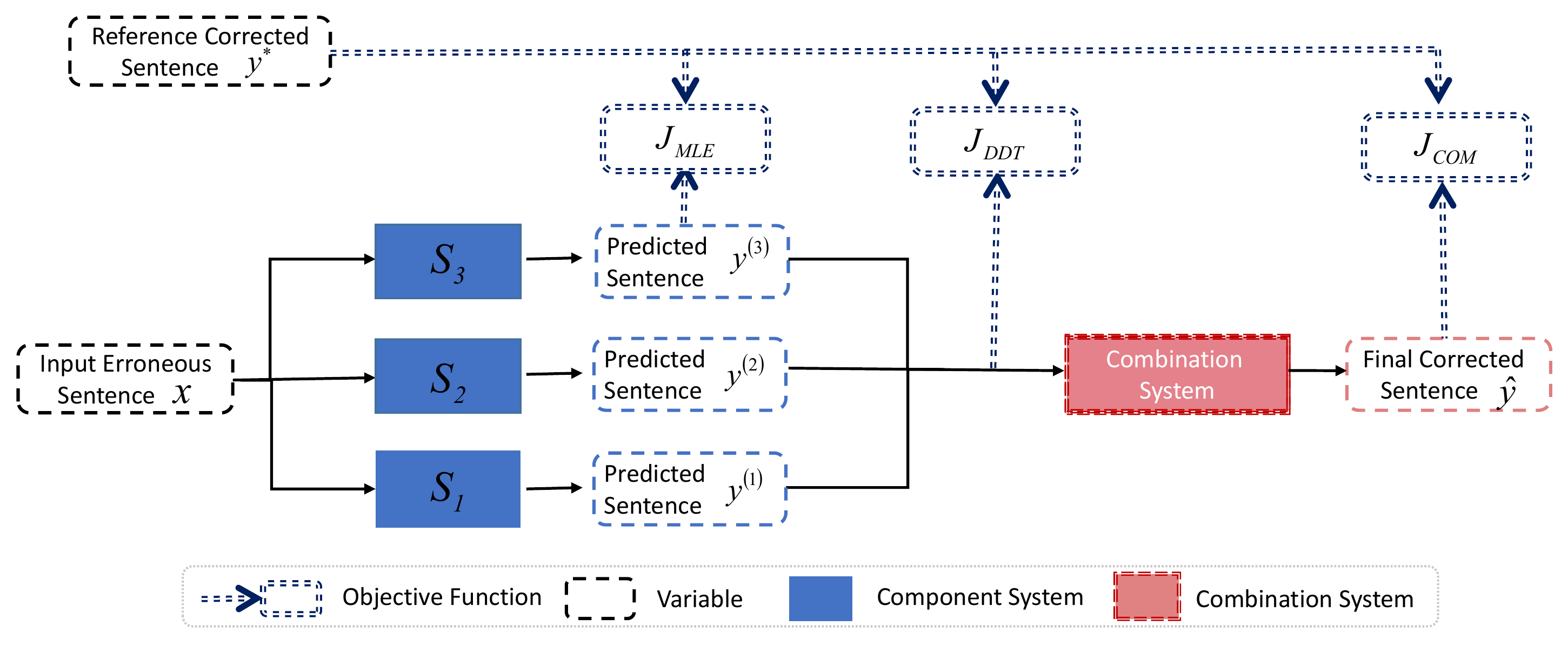}%left down right up 
      \caption{Model illustration. The whole model consists of two parts: \textcolor{blue}{component systems} and \textcolor{red}{combination system}.}
      \label{pic:framework_illus}
      \vspace{-0.5cm}
      \end{center}
      \end{figure*}

DDC consists of two parts (i.e., \textcolor{blue}{component systems} and \textcolor{red}{combination system}) and includes three stages for training (i.e., \textbf{individual training}, \textbf{diversity-driven training}, and \textbf{training of the combination system}). Before combination, the first two stages, individual training and diversity-driven training, are performed and the backbone component is trained to produce both diverse and accurate outputs. Then during system combination, we combine the outputs from the diverse component systems to produce better outputs. We illustrate the whole model in Fig. \ref{pic:framework_illus} and explain the two parts and three stages in this section. 

\subsection{Component Systems}
The best performing combination algorithm relies on multiple distinct and equally good component systems. To achieve this, we split the learning of component systems into two stages: individual training to improve performance, and diversity-driven training to enhance diversity besides performance. 

For ease and clarity of explanation, we will illustrate our approach using three component systems $S_1$, $S_2$, and $S_3$ (although our approach is applicable to any number of component systems). Among them, $S_3$ is the selected backbone component that is trainable. $\Phi$ is the set of parameters of $S_3$. The remaining two component systems are treated as black boxes and we only require their final outputs to be available. Given an input erroneous sentence $\mathbf{x}$, we denote the corresponding predicted outputs from the three component systems as $\mathbf{y^{(1)}}$, $\mathbf{y^{(2)}}$, and $\mathbf{y^{(3)}}$, respectively. 

\paragraph{Individual Training} Pairs of erroneous and corrected sentences are used to train the backbone component, namely the trainable system $S_3$. Given an erroneous sentence $\mathbf{x}=(\mathbf{x}_1,\mathbf{x}_2,...,\mathbf{x}_n)$ with $n$ tokens, $\mathbf{y^{*}}=(\mathbf{y}_1^{*},\mathbf{y}_2^{*},...,\mathbf{y}_m^{*})$ denotes the reference corrected sentence of $\mathbf{x}$ with $m$ tokens. As shown in \eqref{eq:0}, training the backbone system typically uses maximum likelihood estimation (MLE), where $\mathcal{X}$ is the set of erroneous sentences and $J_{MLE}$ is the objective function:
\begin{equation} \label{eq:0}
J_{MLE}(\Phi) =  \sum_{\mathbf{x} \in \mathcal{X}} \sum_{t=1}^{m} \log P(\mathbf{y^{*}_{t}}|\mathbf{y^{*}_{<t}},\mathbf{x};\Phi)
\end{equation}
\paragraph{Diversity-Driven Training} 
After the training procedure to learn individual systems following \eqref{eq:0}, we propose a module that can keep the original architecture of the model with good performance and simultaneously reward the model for producing sentences that are different from the outputs of the other components. Since the smallest change is desired, an additional objective function is used to induce diversity among components.

Specifically, to learn $S_3$, we define an objective function that includes the MLE objective in \eqref{eq:0} and the new objective $J_{RL}(\Phi)$ as follows:
    \begin{equation} \label{eq:1}
    \begin{split}
& J_{DDT}(\Phi) \\
=& (1-\alpha) J_{MLE}(\Phi) + \alpha J_{RL}(\Phi) \\
=&  (1-\alpha) \sum_{\mathbf{x} \in \mathcal{X}} \sum_{t=1}^{m} \log P(\mathbf{y^{*}_{t}}|\mathbf{y^{*}_{<t}},\mathbf{x};\Phi) +
 \\
& \alpha \sum_{\mathbf{x} \in \mathcal{X}} E_{\mathbf{y^{(3)}} \sim P(\mathbf{y}^{(3)}|\mathbf{x};\Phi)} R(\mathbf{y^{(1)}},\mathbf{y^{(2)}},\mathbf{y^{(3)}})
    \end{split}
    \end{equation}
where $\alpha$ is a tunable hyperparameter. We use $\alpha$ to represent a trade-off between two criteria: diversity and high accuracy. $\alpha$ ranges from 0 to 1. 

For the second term of \eqref{eq:1}, we define $R(\mathbf{y^{(1)}}, \mathbf{y^{(2)}}, \mathbf{y^{(3)}})$ to measure how different the output of the backbone component system $S_3$ is from the outputs of the other component systems $S_1$ and $S_2$. The larger the difference, the larger the value of $R$. Specifically,
\begin{equation} \label{eq:3}
R(\mathbf{y^{(1)}}, \mathbf{y^{(2)}}, \mathbf{y^{(3)}}) = g(\mathbf{y^{(1)}}, \mathbf{y^{(3)}}) + g(\mathbf{y^{(2)}}, \mathbf{y^{(3)}})
\end{equation}
where $g(\cdot)$ is the minimum edit distance of the two sentences. Note that because $S_1$ and $S_2$ are not trainable, $g(\mathbf{y^{(1)}}, \mathbf{y^{(2)}})$ is not considered.

With the designed objective function, we then consider how to maximize it. For the first term of $J_{DDT}(\Phi)$, we use supervised training to maximize the likelihood of reference sentences. However, $R$ is computed at the sentence level, since it is obtained after a component system generates a complete sentence. Therefore, it is impossible to directly compute the gradient of $\Phi$ for the second term of $J_{DDT}(\Phi)$. Inspired by Wu et al. \cite{wu2018study}, we use the REINFORCE algorithm of Williams \cite{williams1992simple} to solve the problem. Specifically, the backbone component system can be viewed as an agent, while the environment is defined as the previously generated tokens $\mathbf{y}^{(3)}_{<t}$ at each token position $t$ as well as $\mathbf{x}$.
The prediction of each token is an action that the agent picks up from the whole vocabulary. The set of parameters $\Phi$ defines the policy $P(\mathbf{y}^{(3)}_{t}|\mathbf{y}^{(3)}_{<t},\mathbf{x};\Phi)$ that the backbone system obeys to choose an action. The reward $R(\mathbf{y^{(1)}}, \mathbf{y^{(2)}}, \mathbf{y^{(3)}})$ is computed by the outputs of three component systems. The REINFORCE algorithm aims to maximize the expected reward by approximating the expectation via sampling $\mathbf{y}^{(3)}$ from $P(\mathbf{y}^{(3)}|\mathbf{x};\Phi)$. Thus, we optimize the objective  $J_{RL}(\Phi)$ by maximizing $\hat{J}_{RL}(\Phi)$:
\begin{equation} \label{eq:4}
\hat{J}_{RL}(\Phi) = \sum_{\mathbf{x} \in \mathcal{X}, \mathbf{y^{(3)}} \sim P(\mathbf{y}^{(3)}|\mathbf{x};\Phi)}  R(\mathbf{y^{(1)}},\mathbf{y^{(2)}},\mathbf{y^{(3)}})
\end{equation} 
The basic REINFORCE algorithm uses the distribution $P(\mathbf{y}^{(3)}|\mathbf{x};\Phi)$ predicted by $S_3$ to sample once to obtain $\mathbf{y^{(3)}}$ and compute the reward. It is unstable, inefficient, and has high variance. 
As such, we use two methods to stabilize the training process.
The first method is multinomial sampling \cite{chatterjee2010minimum}, which produces each token one
by one through multinomial sampling over the distribution $P(\mathbf{y}^{(3)}|\mathbf{x};\Phi)$.
The second method is baseline reward \cite{weaver2001optimal}, which aims to reduce the variance of the gradient. For baseline reward, we subtract an average reward $\overline{R}$ from multiple samples and the actual
reward that is used to update the policy is $R(\mathbf{y^{(1)}}, \mathbf{y^{(2)}}, \mathbf{y^{(3)}})-\overline{R}$ instead of $R(\mathbf{y^{(1)}}, \mathbf{y^{(2)}}, \mathbf{y^{(3)}})$.
The gradient of $J_{RL}(\Phi)$ in \eqref{eq:1} is computed as follows:
\begin{equation} \label{eq:5}
\begin{split}
\big(R(\mathbf{y^{(1)}}, \mathbf{y^{(2)}}, \mathbf{y^{(3)}})-\overline{R}\big)\nabla_\Phi \log P(\mathbf{y}^{(3)}|\mathbf{x};\Phi)
\end{split}
\end{equation}
For simplicity, in the above description, we assume that there is only one trainable backbone system, namely $S_3$. When multiple component systems are trainable, they take turns to be the backbone component system and update its parameters. For example, when three component systems (i.e., $S_1$, $S_2$, and $S_3$) are all trainable, we first complete diversity-driven training by using $S_1$ as the backbone system while fixing the parameters of $S_2$ and $S_3$. Then we use $S_2$ as the backbone system to update its parameters, while fixing the trained $S_1$ and $S_3$. Lastly, we optimize $S_3$ as the backbone system while fixing the other two trained components.

\subsection{Combination System}\label{sec:cs}
After training the individual component systems, the final component systems can be combined using some system combination scheme.
We do not set constraints on the specific approach of the combination system. As such, our framework can be easily applied to other combination systems without changing their original architectures.

Considering that the lack of in-domain labeled data is a bottleneck for GEC, we would like the system combination approach to be data-efficient and quick to train. Since only several thousand output sentences have been released for current state-of-the-art (SOTA) systems used as component systems, it is advantageous to have a combination system that can be trained with a modest number of training sentences and parameters. Following Susanto et al. \cite{susanto2014system}, we use MEMT, a Multi-Engine Machine Translation system \cite{heafield2010combining}, to combine multiple component systems to improve the overall performance. 
This system combination method has two parts. (1) Creating a search graph: Corrected sentences (also called hypotheses) from component systems are aligned using the METEOR aligner \cite{banerjee2005meteor}. A resulting search space \cite{heafield2009machine} is defined on aligned words related to each word.
(2) Beam search over the graph: Beam search is used from left to right to generate the final sentence word-by-word and the search actions are scored using a linear model that combines a set of defined features. The trainable parameters are the weights of the features. Model parameters are tuned to optimize its objective $J_{COM}$ using Z-MERT \cite{zaidan2009z}. 

\section{Experiments}

\subsection{Data}
\begin{table*}[]
\centering
\caption{Statistics of the synthetic training data, human-corrected training data, development data, and test data. \textit{NUS}: 25 students at the National University of Singapore (NUS) wrote new essays to serve as the blind test data in the CoNLL-2014 shared task. ``1'' in the \textbf{Oversampled} column means the corpus has not been oversampled.}
\resizebox{0.95\textwidth}{!}{
\begin{tabular}{r|c|c|c|c|c}
\hline
\textbf{}                                           & \textbf{Type}         & \textbf{Source}       & \textbf{Oversampled} & \textbf{No. of Samples} & \textbf{No. of Source Tokens} \\ \hline
\multirow{5}{*}{\textbf{Synthetic Training Data}} & \multirow{4}{*}{Type 1} & \textit{Gutenberg}    & 1                    & 11.60M                     & 240.17M                       \\ \cline{3-6} 
                                                    &                         & \textit{Tatoeba}      & 1                    & 1.17M                     & 10.89M                        \\ \cline{3-6} 
                                                    &                         & \textit{WikiText-103} & 1                    & 3.93M                     & 97.22M                        \\ \cline{3-6} 
                                                    &                         & \textit{News Crawl}   & 1                    & 33.30M                    & 754.20M                       \\ \cline{2-6} 
                                                    & Type 2                  & \textit{News Crawl}   & 1                    & 100.00M                   & 2775.24M                      \\ \hline
\multirow{4}{*}{\textbf{Human-corrected Training Data}}    & \multirow{4}{*}{-} & \textit{FCE}          & 10                   & 30,541                    & 489,886                       \\ \cline{3-6} 
                                                    &                         & \textit{NUCLE}        & 5                    & 57,151                    & 1,161,719                         \\ \cline{3-6} 
                                                    &                         & \textit{W\&I Train}   & 10                   & 34,308                    & 628,720                       \\ \cline{3-6} 
                                                    &                         & \textit{Lang-8}       & 1                    & 1,114,139                     & 12,948,632                        \\ \hline
\textbf{BEA-2019 Dev Data}                           & -                  & \textit{W\&I + LOCNESS Dev}    & 1                    & 4,384                      & 86,973                        \\ \hline
\textbf{CoNLL-2014 Test Data}                               & -                  & \textit{NUS}   & 1                    & 1,312                      &         30,144                \\ \hline
\textbf{BEA-2019 Test Data}                               & -                  & \textit{W\&I + LOCNESS Test}   & 1                    & 4,477                      & 85,668                        \\ \hline
\end{tabular}%
}
\vspace{-0.5cm}
\label{tab:data_analysis}
\end{table*}

We collect all data from the BEA-2019 shared task\footnote{\url{https://www.cl.cam.ac.uk/research/nl/bea2019st/}} \cite{bryant2019bea} and the CoNLL-2014 shared task\footnote{\url{https://www.comp.nus.edu.sg/~nlp/conll14st/conll14st-test-data.tar.gz}} \cite{ng2014conll}.
For both shared tasks, we use the same data for training and validation.
We list the data used for training, validation, and testing as follows:

\textbf{Synthetic training data: }Four publicly available corpora are used for model pretraining: the Gutenberg dataset \cite{lahiri2014complexity}, the Tatoeba dataset\footnote{\url{https://tatoeba.org/eng/downloads}}, the WikiText-103 dataset \cite{merity2016pointer}, and the News Crawl dataset \cite{chollampatt2018multilayer}. Two types of training data as indicated in Table \ref{tab:data_analysis} are used to build ensemble of component systems. Synthetic training data is a collection of \textit{(perturbed, correct)} sentence pairs, where \textit{correct} is a grammatically correct sentence from these four corpora, while \textit{perturbed} is the corresponding sentence obtained from \textit{correct} by introducing common grammatical errors and character errors following the approach of Choe et al. \cite{choe2019neural} and Kiyono et al. \cite{kiyono2019empirical}.

\textbf{Human-corrected training data: }Parallel training data is a collection of \textit{(erroneous, correct)} sentence pairs. \textit{erroneous} is a sentence written by native or non-native English speakers. \textit{correct} is produced by a human teacher after correcting errors in \textit{erroneous}. Parallel training data provided in the BEA-2019 shared task is from four sources: the FCE corpus \cite{yannakoudakis2011new}, W\&I+LOCNESS datasets \cite{yannakoudakis2018developing}, NUCLE \cite{dahlmeier2013building}, and a pre-processed version of the Lang-8 Corpus of Learner English \cite{mizumoto2011mining}. Since the number of parallel training sentences is limited, the parallel training data is built by oversampling some corpora, where the number of times a corpus is oversampled is indicated in Table \ref{tab:data_analysis}.

\textbf{Development data: }W\&I+LOCNESS Dev data is the development dataset.
Note that the same development dataset is used in both the BEA-2019 and CoNLL-2014 shared tasks in this paper, since we are only able to obtain the outputs of the component systems used in this paper on the W\&I+LOCNESS Dev data.

\textbf{Test data: }
There are two test datasets, one for the BEA-2019 shared task and the other for the CoNLL-2014 shared task. We report results in the restricted setting of training using publicly available training datasets only.

\subsection{Setup}
\paragraph{Setup of Backbone Component}
The backbone component system is required to be trainable to induce information from other systems.
We use the \textbf{T}ransformer-\textbf{B}ig architecture \cite{vaswani2017attention}.
We share the embeddings of the output layer, decoder, and encoder. We name this model \textbf{TB}. We also train an ensemble of four TB models to obtain \textbf{ENS-TB}. These four models are independently trained with different random seeds (two models trained using Type 1 data and two models trained using Type 2 data in Table \ref{tab:data_analysis}). The ensemble model then aggregates the prediction of each individual model, and weights them equally to obtain the final prediction. We summarize the performance of the backbone components in Table \ref{tab:component}. 

\paragraph{Setup of Other Component Systems}
To achieve successful system combination, the component systems need to be of comparable quality. According to this principle, we include several systems that are available and obtain their corrected sentences predicted by their trained models. We summarize the performance of the selected component systems in Table \ref{tab:component}. 

\begin{table}[]

\centering
\caption{Component systems for the BEA-2019 shared task and CoNLL-2014 shared task. -: Not available.}
\resizebox{0.45\textwidth}{!}{%
\begin{tabular}{cccc}
\hline
\multicolumn{1}{c|}{\textbf{Component Systems}}                          & \multicolumn{1}{c|}{\textbf{Prec.}} & \multicolumn{1}{c|}{\textbf{Rec.}} & \textbf{$F_{0.5}$} \\ \hline

\multicolumn{4}{c}{\textit{\small{BEA-2019 Test (ERRANT)}}}                                                                                                                \\ \hline
\multicolumn{1}{l|}{Choe et al. \cite{choe2019neural}} & \multicolumn{1}{c|}{75.19}          &   \multicolumn{1}{c|}{51.91}         & 69.00            \\ \hline
\multicolumn{1}{l|}{\multirow{2}{*}{\begin{tabular}[c]{@{}l@{}}Grundkiewicz, Junczys-Dowmunt,\\ and Heafield \cite{grundkiewicz2019neural}\end{tabular}}}            & \multicolumn{1}{c|}{\multirow{2}{*}{\begin{tabular}[c]{@{}c@{}}72.28\end{tabular}}}          & \multicolumn{1}{c|}{\multirow{2}{*}{\begin{tabular}[c]{@{}c@{}}60.12\end{tabular}}}         &  \multicolumn{1}{c}{\multirow{2}{*}{\begin{tabular}[c]{@{}c@{}}69.47\end{tabular}}}        \\ \multicolumn{1}{l|}{}             & \multicolumn{1}{c|}{}          & \multicolumn{1}{c|}{}         &  {}        \\ \hline
\multicolumn{1}{l|}{Kiyono et al \cite{kiyono2019empirical}}                      & \multicolumn{1}{c|}{72.1}           & \multicolumn{1}{c|}{61.8}          & 69.8         \\ \hline
\multicolumn{1}{l|}{ENS-TB}                           & \multicolumn{1}{c|}{73.89}               & \multicolumn{1}{c|}{57.52}              &      69.91         \\ \hline

\multicolumn{4}{c}{\textit{\small{CoNLL-2014 ($M^2$ scorer)}}}                                                                                                                 \\ \hline
\multicolumn{1}{l|}{\multirow{2}{*}{\begin{tabular}[c]{@{}l@{}}Grundkiewicz, Junczys-Dowmunt,\\ and Heafield \cite{grundkiewicz2019neural}\end{tabular}}}            & \multicolumn{1}{c|}{\multirow{2}{*}{\begin{tabular}[c]{@{}c@{}}-\end{tabular}}}          & \multicolumn{1}{c|}{\multirow{2}{*}{\begin{tabular}[c]{@{}c@{}}-\end{tabular}}}         &  \multicolumn{1}{c}{\multirow{2}{*}{\begin{tabular}[c]{@{}c@{}}64.16\end{tabular}}}        \\ 
\multicolumn{1}{l|}{}             & \multicolumn{1}{c|}{}          & \multicolumn{1}{c|}{}         &  {}        \\ \hline

\multicolumn{1}{l|}{\multirow{2}{*}{\begin{tabular}[c]{@{}l@{}}Kiyono et al \cite{kiyono2019empirical}\\ (PRETLARGE+SSE+R2L+SED)\end{tabular}}}            & \multicolumn{1}{c|}{\multirow{2}{*}{\begin{tabular}[c]{@{}c@{}}73.3\end{tabular}}}          & \multicolumn{1}{c|}{\multirow{2}{*}{\begin{tabular}[c]{@{}c@{}}44.2\end{tabular}}}         &  \multicolumn{1}{c}{\multirow{2}{*}{\begin{tabular}[c]{@{}c@{}}64.7\end{tabular}}}        \\ 
\multicolumn{1}{l|}{}             & \multicolumn{1}{c|}{}          & \multicolumn{1}{c|}{}         &  {}        \\ \hline

\multicolumn{1}{l|}{Kaneko et al \cite{kaneko2020encoder}}                      & \multicolumn{1}{c|}{72.6}           & \multicolumn{1}{c|}{46.4}          & 65.2          \\ \hline
\multicolumn{1}{l|}{Omelianchuk et al.\cite{omelianchuk2020gector}}                           & \multicolumn{1}{c|}{78.2}               & \multicolumn{1}{c|}{41.5}              &      66.5         \\ \hline
\multicolumn{1}{l|}{ENS-TB}                           & \multicolumn{1}{c|}{70.96}               & \multicolumn{1}{c|}{43.24}              &     62.89         \\ \hline

\end{tabular}
}

\label{tab:component}
\end{table}

\paragraph{Setup of Individual Training}
The individual training steps for the backbone component are as follows.
\begin{itemize}
\item \textbf{Preprocessing:}
First, a neural spellchecker proposed by Choe et al. \cite{choe2019neural} is used to correct the spelling errors of the input erroneous sentences, similar to what was done by Choe et al. \cite{choe2019neural}.
All data is then tokenized using spaCy \cite{honnibal2017spacy}. Finally, the input sentences are byte-pair encoded (BPE) \cite{sennrich2016neural} before feeding into the backbone component. The maximum length of an input sentence is limited to 100 subword units of BPE.

\item \textbf{Pretraining:} Synthetic sentence pairs of the form \textit{(perturbed, correct)} are used to pretrain the backbone component. 

\item \textbf{Fine-tuning:} Human-corrected sentence pairs of the form \textit{(erroneous, correct)} are used to fine-tune the backbone component. The parameters in the fine-tuning step are initialized with the learned parameters of the pretraining step. 
\end{itemize}

\paragraph{Setup of Diversity-Driven Training}

During diversity-driven training (DDT), we preserve all the hyperparameters from the individual training step. Since most SOTA component systems do not release their predictions on the BEA-2019 and CoNLL-2014 training data, Kantor et al. \cite{kantor2019learning} use the development data in the training stage. For the same reason, we also use the BEA-2019 dev data to train the component systems in the DDT stage. We further design an analysis experiment in Section \ref{sec:analysis_dev} to ensure that our approach does not encounter the overfitting issue in the development dataset. $\alpha$ is set to 0.5.

\paragraph{Evaluation Metrics}
For the CoNLL-2014 shared task, models are evaluated using the MaxMatch scorer \cite{dahlmeier2012better}. The MaxMatch scorer measures performance by span-based correction using $F_{0.5}$, which weights precision twice as much as recall. For the BEA-2019 shared task, we evaluate models using the ERRANT scorer \cite{bryant2019bea}.

\subsection{Effectiveness of Diversity}
\begin{figure}[ht]
      \begin{center}
      \includegraphics[width=1\columnwidth,trim=0 6 0 0,clip]{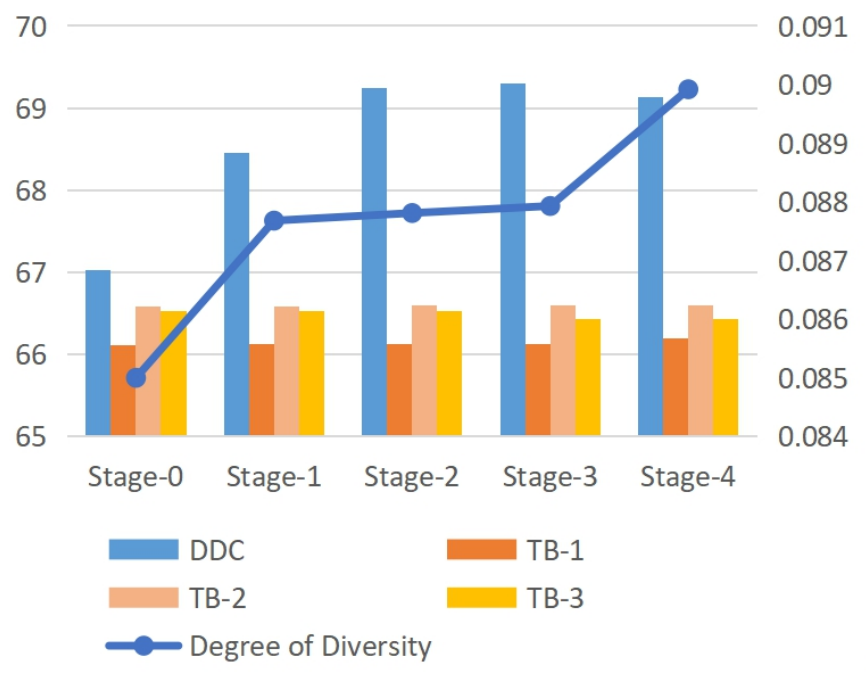}
      \caption{Combination results with versus without encouraging diversity. Component systems take turns to complete diversity-driven training and update the model parameters. Stage-n denotes the n-th diversity-driven training step. Stage-0: Without diversity-driven training (i.e., MEMT).}
      \label{tab:similar-model}
       \end{center}
      \end{figure}

We study the impact of diversity on the performance of system combination. We experiment with combining three TB models with different random seeds (i.e., \textit{TB-1}, \textit{TB-2} and \textit{TB-3}) and report the results on BEA-2019 test data. Since these three models share the same architecture, the corrected sentences from these three models are similar to one another. Without encouraging diversity (``Stage-0'' in Fig. \ref{tab:similar-model}), the performance of combination $F_{0.5}$ (67.03) is similar to the individual components. Note that ``Stage-0'' is actually MEMT.

Then, we design a series of experiments to evaluate the effect of gradually increasing diversity.
First in Stage-1, we perform diversity-driven training on \textit{TB-1} and evaluate the performance of combining \textit{TB-2}, \textit{TB-3}, and the trained \textit{TB-1}. Compared with the baseline Stage-0 (directly combining using MEMT without diversity-driven training) in Fig. \ref{tab:similar-model}, a notable improvement on $F_{0.5}$ is obtained. Then, the component models (\textit{TB-2}, \textit{TB-3}, and \textit{TB-1}) take turns to complete diversity-driven training and result in Stage-2, Stage-3, and Stage-4. 
From Stage-0 to Stage-3, the component systems gradually contain more diversity when measured by $1 - BLEU$, where $BLEU$ is the average BLEU score of two outputs. The results show that the performance of combination consistently benefits from increasing diversity from Stage-0 to Stage-3. But too much diversity, for example in Stage-4, can cause a decrease in performance.

We perform statistical significance tests on $F_{0.5}$ scores of the development data, based on one-tail sign test with bootstrap re-sampling on 100 samples. We compare the outputs of each pair of models: Stage-0, Stage-1, Stage-2, and Stage-3. For each pair, the improvement is statistically significant ($p<0.001$).

\subsection{Results on BEA-2019 Shared Task}
\begin{table*}[htb]
\centering
\caption{Comparison of recent SOTA approaches. The ERRANT scores for the BEA-2019 test data and the BEA-2019 dev data are reported. -: Not available. Approaches with $*$ are used as our component systems.}
\resizebox{0.85\textwidth}{!}{%
\begin{tabular}{l|c|c|c|c|c|c|c}
\hline
\multicolumn{1}{c|}{\multirow{2}{*}{\textbf{Model}}} & \multirow{2}{*}{\textbf{Ensemble}} & \multicolumn{3}{c|}{\textbf{\begin{tabular}[c]{@{}c@{}}BEA-2019 Dev\\ (ERRANT)\end{tabular}}} & \multicolumn{3}{c}{\textbf{\begin{tabular}[c]{@{}c@{}}BEA-2019 Test\\ (ERRANT)\end{tabular}}} \\ \cline{3-8} 
\multicolumn{1}{c|}{}                                &                                    & \textit{Prec.}                 & \textit{Rec.}                 & \textit{$F_{0.5}$}                & \textit{Prec.}                 & \textit{Rec.}                 & \textit{$F_{0.5}$}                 \\ \hline
Chen et al. \cite{chen2020improving} &       N                             & -                         & -                         & -                        & 70.4                          & 55.9                        & 66.9                            \\ \hline

Choe et al. \cite{choe2019neural}$*$  &       Y                             & 62.73                          & 33.23                         & 53.27                        & 75.19                          & 51.91                         & 69.00                            \\ \hline
Grundkiewicz, Junczys-Dowmunt, and Heafield \cite{grundkiewicz2019neural}$*$            &       Y                             & 59.1                           & 36.8                          & 53                           & 72.28                          & \textbf{60.12}                         & 69.47                         \\ \hline
 Kiyono et al. \cite{kiyono2019empirical} (PRETLARGE+SSE+R2L)         & Y                                  &      -                          &            -                   &        -                      & 72.1                           & 61.8                          & 69.8                          \\ \hline
 Kiyono et al. \cite{kiyono2019empirical} (PRETLARGE+SSE+R2L+SED)$*$         & Y                                  &      -                          &            -                   &        -                      & 74.7                           & 56.7                          & 70.2                          \\ \hline
Kantor et al. \cite{kantor2019learning}               & Y                                  & -                              & -                             & -                            & 78.31                          & 58                            & 73.18                         \\ \hline 
Omelianchuk et al.\cite{omelianchuk2020gector} (BERT + RoBERTa + XLNet)                      & Y                                  & -                              & -                             & -                            & 78.9                           & 58.2                          & 73.6                          \\ \hline
MEMT                        & Y                                  &         65.26	                       &            \textbf{38.68}                 &              57.38                & 78.13                          & 60.11                         & 73.71                         \\ \hline \hline
DDC                          & Y                                  &      \textbf{66.54}                          &         \textbf{37.90}                      &            \textbf{57.81}                  & \textbf{80.25}                          & 58.91                         & \textbf{74.83}                         \\ \hline

\end{tabular}
}

\label{tab:bea-table}
\end{table*}

In Table \ref{tab:bea-table}, we summarize the results over the BEA-2019 shared task, the most recent GEC shared task.
Our model consistently outperforms the component systems. The improvement achieved is statistically significant ($p < 0.001$), based on one-tail sign test with bootstrap resampling.

\begin{table}[htb]
\centering
\caption{Comparison of recent SOTA approaches. The $M^2$ scores for CoNLL-2014 test data are reported. -: Not available. Approaches with $*$ are used as our component systems.}
\resizebox{0.47\textwidth}{!}{%
\begin{tabular}{l|c|c|c}
\hline
\multicolumn{1}{c|}{\multirow{2}{*}{\textbf{Model}}}  & \multicolumn{3}{c}{\textbf{\begin{tabular}[c]{@{}c@{}}CoNLL-2014 Test\\ ($M^2$ scorer)\end{tabular}}} \\ \cline{2-4} 
\multicolumn{1}{c|}{}                                & \textit{Prec.}                 & \textit{Rec.}                 & \textit{$F_{0.5}$}                 \\ \hline
 
Wang and Zheng \cite{wang2020improving}                                                                     & 62.9                          & 39.9                          & 56.4                         \\ \hline

Choe et al. \cite{choe2019neural}                                                                     & 74.76                          & 34.05                         & 60.33                         \\ \hline

Lichtarge et al. \cite{lichtarge2019corpora}                                                                    & 66.7                           & 43.9                          & 60.4                          \\ \hline

Chen et al. \cite{chen2020improving}                                                                    & 72.6                           & 37.2                          & 61.0                          \\ \hline

Awasthi et al. \cite{awasthi2019parallel}                                                                       & 68.3                           & 43.2                          & 61.2                          \\ \hline

\multicolumn{1}{l|}{\multirow{2}{*}{\begin{tabular}[c]{@{}l@{}}Grundkiewicz, Junczys-Dowmunt,\\ and Heafield \cite{grundkiewicz2019neural}$*$\end{tabular}}}            & \multicolumn{1}{c|}{\multirow{2}{*}{\begin{tabular}[c]{@{}c@{}}-\end{tabular}}}          & \multicolumn{1}{c|}{\multirow{2}{*}{\begin{tabular}[c]{@{}c@{}}-\end{tabular}}}         &  \multicolumn{1}{c}{\multirow{2}{*}{\begin{tabular}[c]{@{}c@{}}64.16\end{tabular}}}        \\ 
\multicolumn{1}{l|}{}             & \multicolumn{1}{c|}{}          & \multicolumn{1}{c|}{}         &  {}        \\ \hline

\multicolumn{1}{l|}{\multirow{2}{*}{\begin{tabular}[c]{@{}l@{}}Kiyono et al \cite{kiyono2019empirical}\\ (PRETLARGE+SSE+R2L+SED)$*$\end{tabular}}}            & \multicolumn{1}{c|}{\multirow{2}{*}{\begin{tabular}[c]{@{}c@{}}73.3\end{tabular}}}          & \multicolumn{1}{c|}{\multirow{2}{*}{\begin{tabular}[c]{@{}c@{}}44.2\end{tabular}}}         &  \multicolumn{1}{c}{\multirow{2}{*}{\begin{tabular}[c]{@{}c@{}}64.7\end{tabular}}}        \\ 
\multicolumn{1}{l|}{}             & \multicolumn{1}{c|}{}          & \multicolumn{1}{c|}{}         &  {}        \\ \hline

\multicolumn{1}{l|}{\multirow{2}{*}{\begin{tabular}[c]{@{}l@{}}Kiyono et al \cite{kiyono2019empirical}\\ (PRETLARGE+SSE+R2L)\end{tabular}}}            & \multicolumn{1}{c|}{\multirow{2}{*}{\begin{tabular}[c]{@{}c@{}}72.4\end{tabular}}}          & \multicolumn{1}{c|}{\multirow{2}{*}{\begin{tabular}[c]{@{}c@{}}46.1\end{tabular}}}         &  \multicolumn{1}{c}{\multirow{2}{*}{\begin{tabular}[c]{@{}c@{}}65.0\end{tabular}}}        \\ 
\multicolumn{1}{l|}{}             & \multicolumn{1}{c|}{}          & \multicolumn{1}{c|}{}         &  {}        \\ \hline

\multicolumn{1}{l|}{Kaneko et al \cite{kaneko2020encoder}$*$}                     & \multicolumn{1}{c|}{72.6}           & \multicolumn{1}{c|}{\textbf{46.4}}          & 65.2          \\ \hline
Omelianchuk et al.\cite{omelianchuk2020gector} $*$                                                       & 78.2                          & 41.5                          & 66.5   \\ \hline

MEMT                                                                                   &            76.39                         &      43.12                        &   66.18                \\ \hline \hline
DDC                                                   &     \textbf{76.43}                          &      45.54                        &   \textbf{67.30} 
\\ \hline 

\end{tabular}
}

\label{tab:conll-table}
\end{table}

\subsection{Results on CoNLL-2014 Shared Task}
We summarize the results on the CoNLL-2014 test data in Table \ref{tab:conll-table}. DDC achieves a performance gain over the component systems and outperforms other approaches. The significance test shows that the improvement is statistically significant.

Recently, Stahlberg and Kumar \cite{stahlberg2021synthetic} obtained $F_{0.5}$ scores of $74.9$ and $68.3$ for BEA-2019 and CoNLL-2014 respectively. They used an external annotation tool and an extremely large synthetic data set with 200 million sentences. In contrast, our approach can boost the weak models to a performance that is close to Stahlberg and Kumar \cite{stahlberg2021synthetic}. For example we boost $F_{0.5}$ from the average score of $69.42$ to $74.83$ on BEA-2019 test data. Moreover, our approach is data-efficient with only around 5,000 training sentences.

\subsection{Speed Comparison}
We measure the average training time in Table \ref{tab:gec_speed}. In addition to the training time, the number of parameters and the machine used are provided. For pretrainig and fine-tuning, we measure the training time for one single model of ENS-TB.
Training a component model, including pretraining, fine-tuning and reranking, is time-consuming. 
Compared to training a single model (e.g., model from Omelianchuk et al. \cite{omelianchuk2020gector}) from scratch, diversity-driven training (DDT) and system combination (SC) are much more efficient.

\begin{table}[htb]
\centering
\caption{Training time for different stages. DDT: Diversity-Driven Training. SC: System Combination. $\ddag$: Calculated by our reproduction.}
\resizebox{0.49\textwidth}{!}{%
\begin{tabular}{r|c|c|c}
\hline
\textbf{Stage}   & \begin{tabular}[c]{@{}c@{}}\# \textbf{parameters}\\ (Million)\end{tabular} & \begin{tabular}[c]{@{}c@{}} \textbf{Time}\\ (Hour)\end{tabular} & \textbf{Machine} \\ \cline{1-4} 
Pretraining &                   205.03                                                &        102.68                                                       &  Tesla V100    \\ \hline
Fine-tuning &        205.03                                                           &    6.67                                                               &  Tesla V100     \\ \hline
Reranking &             $<$1                                                      &         0.50                                                          &   Tesla V100    \\ \hline
DDT &     205.03                                                              &    0.20                                                               &   Tesla V100    \\ \hline 
 \multicolumn{1}{r|}{\multirow{2}{*}{\begin{tabular}[c|]{@{}c@{}} SC \end{tabular}}}          & \multicolumn{1}{c|}{\multirow{2}{*}{\begin{tabular}[c|]{@{}c@{}} $<$1 \end{tabular}}} &\multicolumn{1}{c|}{\multirow{2}{*}{\begin{tabular}[c|]{@{}c@{}} 0.25 \end{tabular}}}& \multicolumn{1}{c}{\multirow{2}{*}{\begin{tabular}[c]{@{}c@{}}Intel Xeon \\ CPU E5649\end{tabular}}}             \\ \multicolumn{1}{c|}{}             & \multicolumn{1}{c|}{}              &  \multicolumn{1}{c|}{}   &  \multicolumn{1}{c}{}  \\\hline \hline
 Omelianchuk et al.\cite{omelianchuk2020gector} $\ddag$ & $>345$ & $>120$ & Tesla V100 \\ \hline
\end{tabular}
}

\label{tab:gec_speed}
\vspace{-0.5cm}
\end{table}

\section{Analysis}

\subsection{Analysis of Choice of Reward Design}
We have experimented with different setups of the reward: BLEU, the number of different tokens, and the minimum edit distance. The results on the BEA-2019 test dataset are summarized in Table \ref{tab:reward}. DDC with different designs of rewards consistently outperforms MEMT. Among them, minimum edit distance performs the best.

\begin{table}[htb]
\centering
\caption{Comparison of different designs of rewards.}
\resizebox{0.40\textwidth}{!}{%
\begin{tabular}{l|c|c|c}
\hline
\multicolumn{1}{c|}{\multirow{2}{*}{\textbf{Reward Design}}}  & \multicolumn{3}{c}{\textbf{\begin{tabular}[c]{@{}c@{}}BEA-2019 Test\\ (ERRANT)\end{tabular}}} \\ \cline{2-4} 
\multicolumn{1}{c|}{}                                & \textit{Prec.}                 & \textit{Rec.}                 & \textit{$F_{0.5}$}                 \\ \hline
BLEU & 78.62 & 59.74 & 73.95   \\ \hline
$\#$Different tokens & 79.71 & 59.87 & 74.76   \\ \hline
Minimum edit distance & 80.25 & 58.91 & 74.83   \\ \hline

\end{tabular}
}

\label{tab:reward}
%\vspace{-0.5cm}
\end{table}

\subsection{Analysis of Choice of Training Data}\label{sec:analysis_dev}
To evaluate whether our approach will encounter the overfitting issue in the  development dataset, we experiment with the same hyper-parameters and setup in Fig. \ref{tab:similar-model} except for using the W\&I training dataset for diversity-driven training. The similar $F_{0.5}$ of 69.26 compared with Stage-3 (69.30) shows that the choice of data for diversity-driven training does not make much difference.

\subsection{Analysis of Choice of Combination System}\label{sec:analy-cs}
As mentioned in Section \ref{sec:cs}, we do not set constraints on the specific approach of the combination system. We follow the setup in Fig. \ref{tab:similar-model} and conduct DDC experiments by replacing MEMT with an average ensemble \cite{zhou2002ensembling}. We obtain $F_{0.5}$ of 69.10 that is better than the baseline (66.95) of simply using the average ensemble without encouraging diversity. It shows that this framework can be easily and effectively applied to other combination systems.

\subsection{Error Analysis}
We investigate the span-level correction results for each error type to compare models with and without diversity-driven training. We use ERRANT to measure $F_{0.5}$ for each error type. We show the results of MEMT and DDC in Table \ref{tab:error_type}. The only difference between MEMT and DDC is diversity-driven training.
Table \ref{tab:error_type} shows the results on the top 10 frequent error types in BEA-2019 Dev data. It can be seen that for almost all the error types, DDC outperforms MEMT. It indicates that encouraging diversity generally improves the performance independent of the error type.

\begin{table}[htb]
\centering
\caption{$F_{0.5}$ scores of MEMT and DDC across error types, for the top 10 frequent error types in the BEA-2019 dev data.}
\resizebox{0.27\textwidth}{!}{
\begin{tabular}{ccc}
\hline
\textbf{Error Type} & \textbf{MEMT} & \textbf{DDC}  \\ \hline
PUNCT               & 60.32             & \textbf{61.45} \\
OTHER               & 28.91             & \textbf{29.58} \\
DET                 & 61.46             & \textbf{61.53} \\
PREP                & 58.51             & \textbf{58.55} \\
VERB:TENSE          & \textbf{53.69}             & 51.63          \\
SPELL               & 77.9              & \textbf{79.92} \\
VERB                & 36.18             & \textbf{36.88} \\
ORTH                & \textbf{68.63}             & 67.8           \\
NOUN                & 29.61             & \textbf{31.68} \\
NOUN:NUM            & 64.11             & \textbf{64.7}  \\ \hline
\end{tabular}
}

\label{tab:error_type}
%\vspace{-0.5cm}
\end{table}

\section{Related Work}
\subsection{Grammatical Error Correction}
Recent research in grammatical error correction mainly focuses on two approaches: neural machine translation approaches and sequence labeling approaches. Neural machine translation \cite{sennrich2016edinburgh} approaches use erroneous sentences corresponding to source sentences, and correct sentences corresponding to target sentences \cite{bryant2019bea,grundkiewicz2019neural,kiyono2019empirical}.
Sequence labeling approaches have achieved impressive results on this task. Different from neural machine translation approaches, sequence labeling approaches formalize the task as sequence labeling \cite{malmi2019encode,awasthi2019parallel,omelianchuk2020gector} and not sequence generation. In both formulations, transformer-based architectures \cite{vaswani2017attention} have become the dominant preferred models for the task. The homogeneity of architectures leads to a challenge for combining multiple systems, where diverse component systems are favored.

\subsection{Diversity in System Combination}
To achieve successful system combination, multiple component systems are required to be simultaneously diverse and of comparable performance \cite{macherey2007empirical}. Compared with individual component quality, the distinction among component systems has been ignored during system combination.
Some work obtains diverse systems by utilizing different system architectures, hyperparameters, as well as data selection and weighting \cite{xiao2013bagging,denero2010model,macherey2007empirical}. Cer, Manning, and Jurafsky \cite{cer2013positive} propose to actively generate diverse components by using online-PRO \cite{hopkins2011tuning}. 
Different from their work, we use reinforcement learning to induce diversity into components that can directly compute the gradient of the parameters. Additionally, our strategy needs only a small number of samples for DDT and training combination system, which is suitable for GEC where the lack of in-domain labeled data is a bottleneck.

\section{Conclusion}

We propose Diversity-Driven Combination for GEC, a system combination strategy to encourage diversity among component systems. 
On both CoNLL-2014 shared task and BEA-2019 shared task, our strategy outperforms component systems by a significant margin. 
In addition to the good performance, our strategy is efficient, where only less than 5,000 samples are used for DDT and training MEMT. 

\section*{Acknowledgements}

This research is supported by the National Research Foundation, Singapore under its AI Singapore Programme (AISG Award No: AISG-RP-2019-014). The computational work for this article was partially performed on resources of the National Supercomputing Centre, Singapore (https://www.nscc.sg). We thank Weiqi Wang and Yang Song for training the backbone component system, and Muhammad Reza Qorib for helpful feedback and comments.

\bibliographystyle{IEEEtran}
\bibliography{acl2021}

\end{document}